%% file: EuMW_Paper_LaTeX_Template_A4_V4.tex
\newcommand{\CLASSINPUTtoptextmargin}{19mm}%
\newcommand{\CLASSINPUTbottomtextmargin}{43mm}%
\newcommand{\CLASSINPUTinnersidemargin}{12.9mm}%
\newcommand{\CLASSINPUToutersidemargin}{12.9mm}%
\documentclass[conference,10pt,a4paper]{IEEEtran}
\usepackage{amsmath}
\usepackage{times}
\usepackage{graphicx}
\usepackage{multirow}
\usepackage[none]{hyphenat}
\usepackage{float}
\usepackage{subfig}
\input{EuMW_modify_IEEEtran_18b_CTAN_V4}

\begin{document}
\raggedbottom
%
%
%
\title{mm-Wave Radar Hand Shape Classification Using Deformable Transformers}
%
%
\author{%
\IEEEauthorblockN{%
Athmanarayanan Lakshmi Narayanan\EuMWauthorrefmark{\#1},
Asma Beevi K. T.\EuMWauthorrefmark{\#2}, 
Haoyang Wu\EuMWauthorrefmark{*\#3},
Jingyi Ma\EuMWauthorrefmark{*\#4}, 
W. Margaret Huang\EuMWauthorrefmark{\#5}, 
}
\IEEEauthorblockA{%
\EuMWauthorrefmark{\#}Intel Labs, Santa Clara CA, USA\\
\EuMWauthorrefmark{*}Intel, Intel Labs, Intel Labs China, China\\
\{
\EuMWauthorrefmark{1}athma.lakshmi.narayanan, 
\EuMWauthorrefmark{2}asma.kuriparambil.thekkumpate,
\EuMWauthorrefmark{3}haoyang.wu,
\EuMWauthorrefmark{4}jingyi.ma,
\EuMWauthorrefmark{5}margaret.huang\}@intel.com\\
}
}
%
\maketitle
%
%
\begin{abstract}
A novel, real-time, mm-Wave radar-based static hand shape classification algorithm and implementation are proposed. The method finds several applications in low cost and privacy sensitive touchless control technology using 60 Ghz radar as the sensor input. As opposed to prior Range-Doppler image based 2D classification solutions, our method converts raw radar data to 3D sparse cartesian point clouds.The demonstrated 3D radar neural network model using deformable transformers significantly surpasses the performance results set by prior methods which either utilize custom signal processing or apply generic convolutional techniques on Range-Doppler FFT images. Experiments are performed on an internally collected dataset using an off-the-shelf radar sensor.
\end{abstract}
%
%

\section{Introduction}

Users are demanding for more and more touchless interfaces and controls especially since the onset of COVID pandemic.  Use of mm-wave radar for non-verbal gesture interaction(\cite{zhang2017doppler}, \cite{lien2016soli}) have the advantage of low-cost implementations without the privacy concern. Radar also helps to detect objects in occluded or variable lighting conditions. Radar signal processing extracts the range, speed and angle information of surrounding moving targets. Most radar based indoor solutions mainly focus on the human action or gesture classification. While static gesture (hand shape/pose) recognition has steadily garnered attention, still few questions remain. 

Prior solution for static hand shape recognition \cite{wu2021scalable} uses Region-Of-Interest (ROI) Range-Doppler (RD) data, with custom Constant False Alarm Rate (CFAR) algorithm to reduce background noise followed by 2-dimensional (2D) convolutional neural networks (CNNs) to classify the hand shape. However, this solution can only detect hand shapes at fixed range (30-60 cm) and orientation (hand front facing radar). Challenges of false positives (such as face classified as hand) and limited shape options (finger, palm, fist) also need to be addressed. Moreover, these solutions have difficulties in extending to additional complex hand shapes and similar but distinct hand poses.

As a closer step towards a complete touchless gesture solution, a flexible hand shape/pose recognition using a 60 GHz FMCW radar is proposed without the need for adaptive CFAR or background noise clutter removal. Our solution converts RD maps to 3-dimensional (3D) point cloud (PC) representation and uses 3D deep neural networks (DNNs) to classify the shape using sparse set of points. Such a point cloud representation allows for:

\begin{itemize}
\item	Better utilization of 3D geometry for complex shapes
\item   Leverage the mature techniques and breakthroughs in the much larger field of 3D computer vision
\item   Rotation invariant augmentation during training.
\end{itemize}

Furthermore, we apply state of the art deep learning architectures like deformable vision transformers to the problem of radar point cloud classification enabling self-attention and increasing the scope of re-usability. 

This paper is organized as follows: The problem description and the data set details are given in Section II. In Section III, three different approaches towards radar shape classification are presented. The model performance and experiment results are summarized and benchmarked in Section IV. Finally, Section V. concludes this work.

\begin{figure}[H]
\centering
\includegraphics[width=90mm,height=40mm]{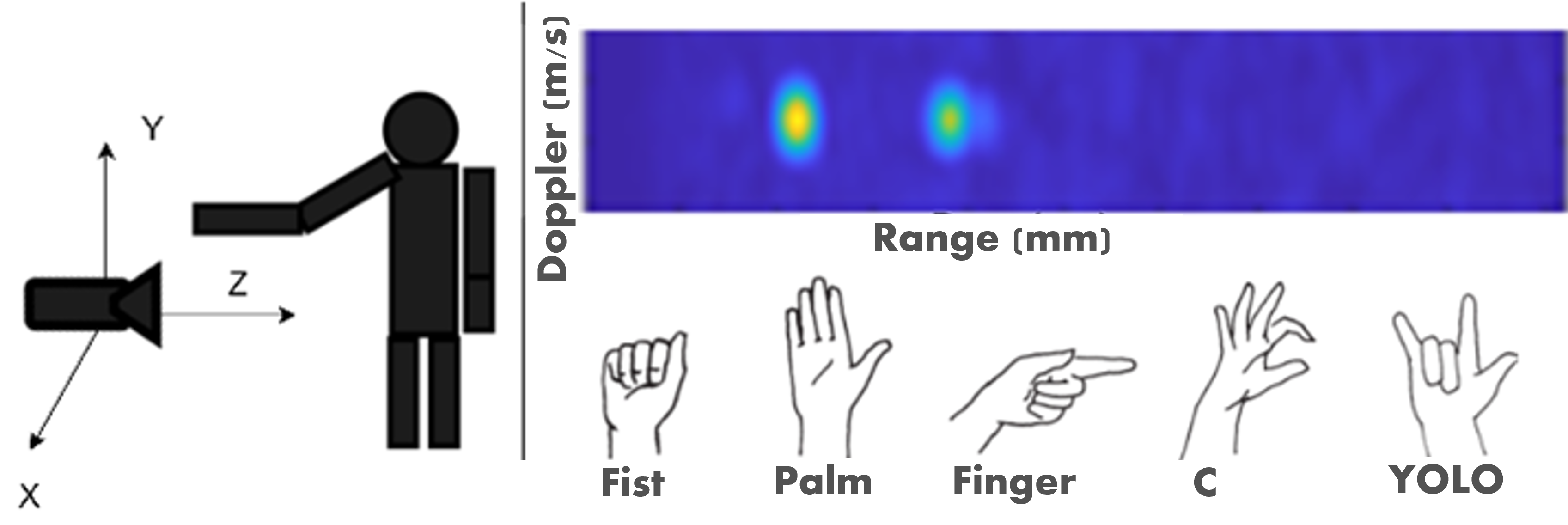}
\caption{ (left) Radar data collection setup and (right) hand pose shapes.}
\label{fig:data_collection}
\end{figure}
\vspace{-\baselineskip}

\section{Problem Description}


\subsection{Hardware setup and data collection}

A low cost, low form factor off-the-shelf radar sensor board (FMCW 60 Ghz radar sensing solution with 1 TX and 3 RX antennas) was used to capture the radar scenes. The radar bandwidth of 5 GHz and ADC sampling rate of 2 MHz was used to capture the RD information with max range and max velocity of about 2m and 17m/s respectively. The radar scenes were captured in a room using both the left and right hands of the subjects within 20-95 cm from the radar sensor board. 

The radar hand shapes dataset consist of 5 different hand shapes which are palm, fist, finger, ‘C’-shape and ‘Yolo’ hand sign. Images of these hand signs \cite{tennant2002american} are given in Fig. 1. The collected dataset consists of about 600 radar data scenarios of two people posing with the hand shapes mentioned above.


\subsection{Data Preparation}

The RD FFT maps for ADC data from the three radar receivers were calculated. The three receiver FFTs were combined to get the azimuth and elevation angle mappings of the radar scene. Radar specific custom and problem tailored signal processing techniques were not used in the data processing pipeline.  

The 3D spatial point clouds are created from the range-FFT maps with amplitude and angle information. Each bin in the 2D FFT map is a point in the polar coordinates having features range, azimuth, elevation, amplitude and doppler. We use spherical to cartesian conversion to convert these FFT maps to point cloud data. The radar range-doppler FFT and radar point cloud data for the hand sign classification dataset is visualized in Fig. 2.

\begin{figure}[H]
\centering
\includegraphics[width=90mm,height=55mm]{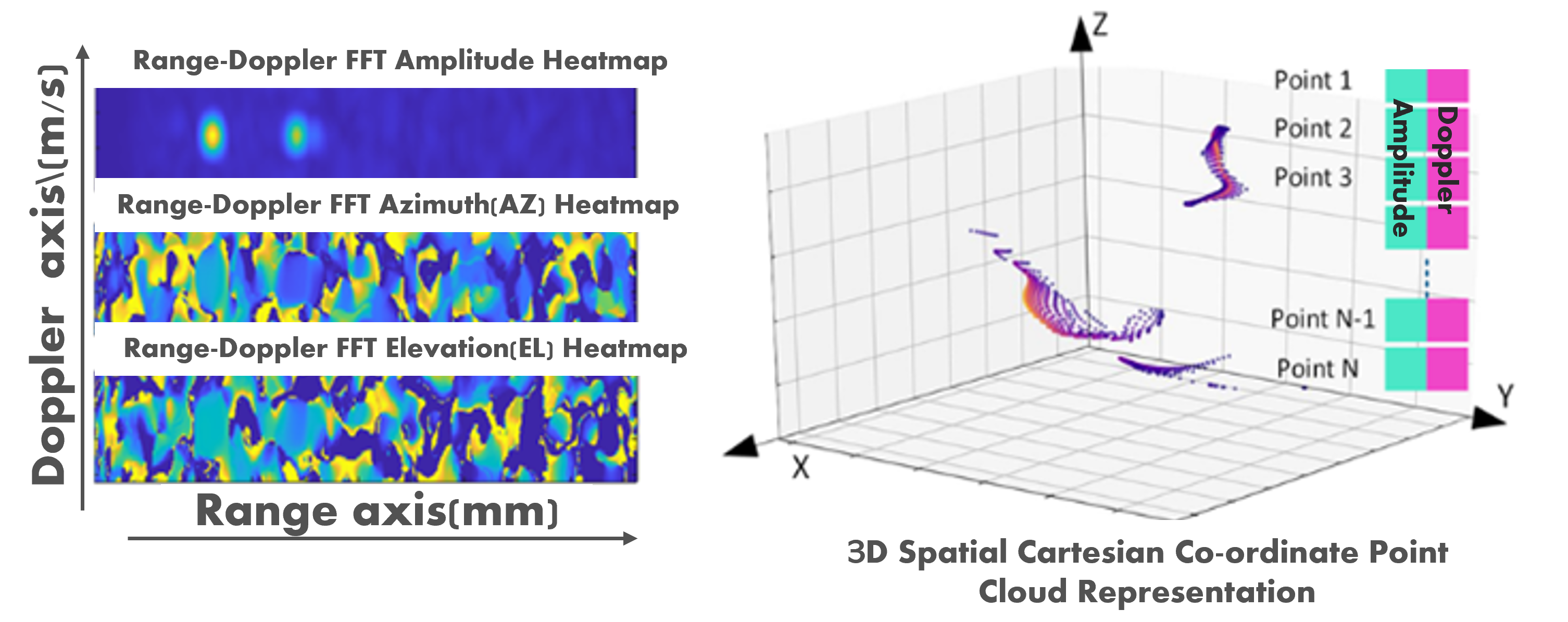}
\caption{(left) Radar RD FFT representation. We visualize the AZ, EL in radians for each bin. (right) Point cloud representation.}
\label{fig:data_collection}
\end{figure}
\vspace{-\baselineskip}

\section{Static Shape Classification}
\label{sec:page style}


\subsection{Introduction}

In this paper, we focus on static hand shape recognition with radar point cloud. While radar point cloud 3D DNNs for automotive use case been demonstrated before\cite{feng2019point}, this work addresses more challenging small static object (hand) classification at close proximity to large objects (such as human body) in indoor noisy scenarios. Input to our DNN is a full resolution radar snapshot of a single human in a room displaying a single hand sign.

While the most common approach is to use the radar images as 2D input to convolutional layers, we hypothesize that using the 3D sparse structure embedding provides a better representation for the DNN models. Hence instead of using Euclidean 2D representation we use irregular cartesian point cloud and compare the 2D and 3D models. To enforce attention on small relevant region in the scene, we also develop a transformer-based solution that further improves the accuracy. In this section, we shall explain the three model architectures developed for comparison, as shown in  Fig. 3.


\subsection{Model Summary}
\label{sec:headings}


\subsubsection{2D CNN Model}
A common input signal for deep learning-based radar perception is to treat the output of 2D-FFT (Range- Doppler maps) on a 3D radar cube (chirps, sampling rate, TX/RX pairs) as a conventional 2D image \cite{abdu2021application}. This 2D image represents the range and Doppler bins along the columns and rows respectively as shown in  Fig. 2. This representation allows for common image-based convolutional neural networks to be re-purposed for radar. To be a fair comparison with the other 3D DNNs we stack the azimuth and elevation maps along with the range Doppler maps as three-channel images for downstream processing using VGG-16\cite{tammina2019transfer}.

\subsubsection{3D CNN Model}
To better model the inherent sparsity and a 3D structure of the radar data, a cartesian point cloud representation is used as input to the 3D CNN model. In this case each radar point includes the XYZ location, as well as features such as received Amplitude and Doppler, resulting in a 5-dimensional tensor. Such a representation can allow for full utilization of the 3D projection context.

In this model a point pillar-based encoder\cite{lang2019pointpillars} is used to scatter the point clouds to image like tensors to simulate a front-view projection. More specifically, the 3D spatial point cloud is arranged in a grid of pillars in the XY plane after voxelization. The point pillar consists of a single layer feed forward neural network and ReLU\cite{nair2010rectified} activation followed by max-pooling among the points in the same pillar. All the points share the feed forward network parameters in the point pillar. The following output is a dense tensor in the XY plane of size 256x256x64. The point pillar featurization of radar point clouds is shown in  Fig. 3a. 

This simulated front view three-channel tensor is processed with 2D CNNs. To keep the number of parameters similar to the VGG-16, we only use three successive 2D convolutions, max-pooling and ReLU combinations followed by two fully connected layers and finally SoftMax to classify the shape.

\begin{figure*}[t]
\centering
\includegraphics[width=150mm]{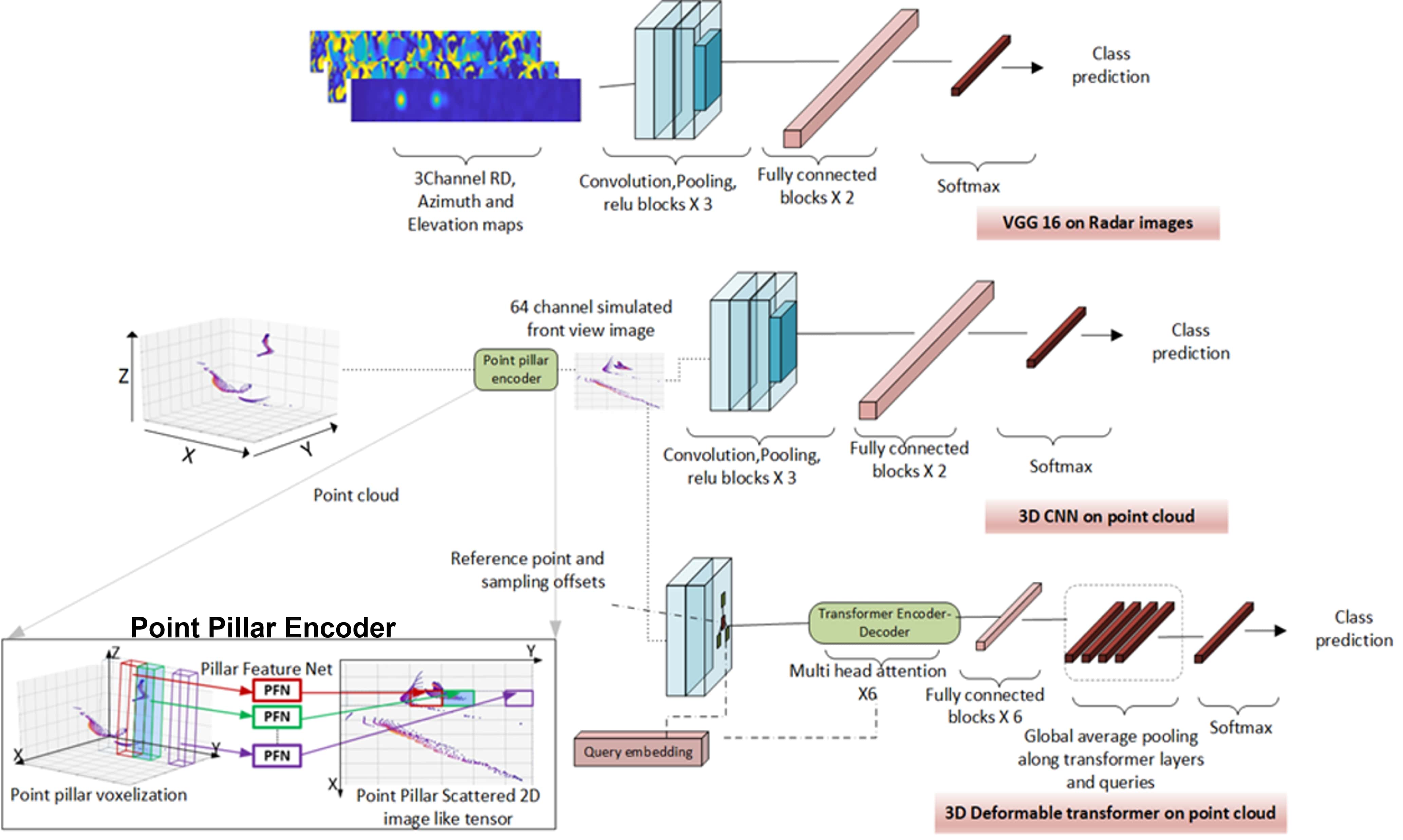}
\caption{A) Point pillar encoder (shown in box) B) Model summary showcasing the inputs and model architecture for VGG-16, 3D CNN and 3D Deformable Transformer  }
\label{fig:bird}
\vspace{-\baselineskip}
\end{figure*}

\subsubsection{3D Deformable Transformer Model}
 The sparse representation contains the full scene (not only the actual hand shape but also the human body as well as the background noise). Hence the model needs to be guided to learn to pay attention to the relevant features and locations and ignore the background information. This is especially crucial is small scale sparse datasets. For example, we do not want the model to learn statistics about the human body location to classify hand shape. 
 
To this regard, a deformable transformer-based model\cite{zhu2020deformable} for the classification of 3D spatial point cloud data is used as backbone. Transformers have been gaining wider adoption not limited to language tasks \cite{dosovitskiy2020image}. Transformer-based model are outperforming CNN alone models in tasks like classification, object detection, semantic segmentation etc.\cite{liu2021swin}. The attention mechanism in transformers allows efficient modelling of pairwise interactions between the elements in the input sequence. However, the transformer models require massive amount of data. The training convergence takes much longer time for vision transformers. Deformable transformers solve this problem by learning to attend to a smaller set of data-dependent locations rather than the entire input by itself. 

Point cloud points are scattered into image like tensors using pillar feature encoding as already described in Section III.B.2.  For the transformer, the 256x256 point pillar output is first fed into convolution layers to down sample to a 42x42 tensor with 256-dimensional feature space which is then subsequently flattened to a 1764x256 tensor.

A sinusoidal positional encoding is added to this tensor to retain the XY spatial information. Six deformable transformer encoder-decoder layers are used to model the self-attention. The transformer has a hidden state feature dimension of 256. The feedforward layers in the transformer have a 2048 feature dimension. We use a total of 100 queries in the decoder to probe various regions of the image like data output from the transformer encoder. Each, decoder hidden state outputs are fed into separate fully connected layers. Global average pooling is used to aggregate the features along the query and the number of decoder layers to produce a single unified prediction. This is further sent to a SoftMax for classification.

{
\setlength{\tabcolsep}{1mm}%
\newcommand{\CPcolumnonewidth}{78mm}%
\newcommand{\CPcolumntwowidth}{88mm}%
\newcommand{\CPcell}[1]{\hspace{0mm}\rule[-0.3em]{0mm}{1.3em}#1}%
\begin{table*}[t]
\caption{Experimental results on best split comparing VGG-16, 3D CNN and 3D deformable transformer. The best results are bolded 
The validation split contains approximately 20 samples for each class to avoid any class imbalance during training and testing.}
\small
\centering
\begin{tabular}{|c|lllll|lll|}
\hline
\multirow{2}{*}{\textbf{Models}}                & \multicolumn{5}{c|}{\textbf{Class wise F-1 score}}                                                                                                                & \multicolumn{3}{l|}{\textbf{Overall Weighted Average Metrics}}                          \\ \cline{2-9} 
                                                & \multicolumn{1}{l|}{C}             & \multicolumn{1}{l|}{Finger}        & \multicolumn{1}{l|}{Palm}          & \multicolumn{1}{l|}{Fist}          & Yolo          & \multicolumn{1}{l|}{Precision}     & \multicolumn{1}{l|}{Recall}        & F1-Score      \\ \hline
2D VGG-16                                       & \multicolumn{1}{l|}{{0.35}} & \multicolumn{1}{l|}{{0.42}} & \multicolumn{1}{l|}{0.24}          & \multicolumn{1}{l|}{0.07}          & 0.15          & \multicolumn{1}{l|}{0.29}          & \multicolumn{1}{l|}{0.28}          & 0.28          \\ \hline
3D CNN                                          & \multicolumn{1}{l|}{0.34}          & \multicolumn{1}{l|}{0.34}          & \multicolumn{1}{l|}{{0.34}} & \multicolumn{1}{l|}{{0.17}} & {0.2}  & \multicolumn{1}{l|}{{0.3}}  & \multicolumn{1}{l|}{{0.29}} & {0.29} \\ \hline
\multicolumn{1}{|l|}{3D Deformable Transformer} & \multicolumn{1}{l|}{\textbf{0.41}} & \multicolumn{1}{l|}{\textbf{0.42}} & \multicolumn{1}{l|}{\textbf{0.36}} & \multicolumn{1}{l|}{\textbf{0.38}} & \textbf{0.34} & \multicolumn{1}{l|}{\textbf{0.44}} & \multicolumn{1}{l|}{\textbf{0.38}} & \textbf{0.38} \\ \hline
\end{tabular}
\label{tab:wordstyles}
\vspace{-\baselineskip}
\end{table*}
}

\section{Experimental Results and Evaluation}

\subsection{Evaluation Summary}

As stated before, hand shape classification was performed on indoor collected radar data samples. Unlike publicly available large scale radar datasets in automotive setting, our dataset is significantly smaller. To alleviate this issue, we perform 5-fold cross validation, with randomly sampled 80\%-20\% train-validation splits, to ensure no overfitting occurs on each architecture. We use splits with relatively equal number of samples for each class in the validation. To ensure fair comparison between the 2D and 3D models, no data augmentation(such as rotation) is performed on the point cloud in this evaluation. The best split results are showcased in Table 1.

Ablation study is performed on each model and the best model is showcased in this work. Specifically, for 2D CNN resize shape for input to VGG16 in tuned and best shape of 224x224 is reported. For the 3D models voxel size, query size and number of transformer decoder-encoder layers are tuned. The reported 3D models have voxel size of (0.0039, 0.0039,1) along X,Y,Z axes, 100 queries and six transformer encoder-decoder pairs.  In this work, we use F1-score, a popular evaluation metric used to compare different models and weight them with the number of samples in each class.

\subsection{Evaluation Summary}
Overall, we see that while going from 2D CNN to 3D CNN only provides an additional 10\% F1-score improvement, the transformer is better able to attend to the hand location and outperforms the others by 9\%. While providing benefits for all the classes, it is most interestingly able to differentiate between “Yolo” and “Fist” which appear almost similar due to the current radar resolution. This is showcased in the confusion matrices in Fig. 4.

Compared with the prior solution (2D VGG-16)\cite{wu2021scalable}, we collected data over a wider range of distances and angles. From the result, we observed that prior solution has been unable to support more flexible hand position and it’s difficult to expand to more hand shapes/poses changes. In addition, our methods does not require any extra custom computations such as ROI and CFAR. To improve the accuracy, MIMO radar can be potentially used to get better angular resolution resulting in a more descriptive point cloud representation increasing the accuracy. Moreover, it is imperative to increase the dataset size to scale with our model parameters and to further optimize the transformer models to support such small-scale sparse datasets. 


\begin{figure}[H]
\centering
\includegraphics[width=90mm,height=140mm]{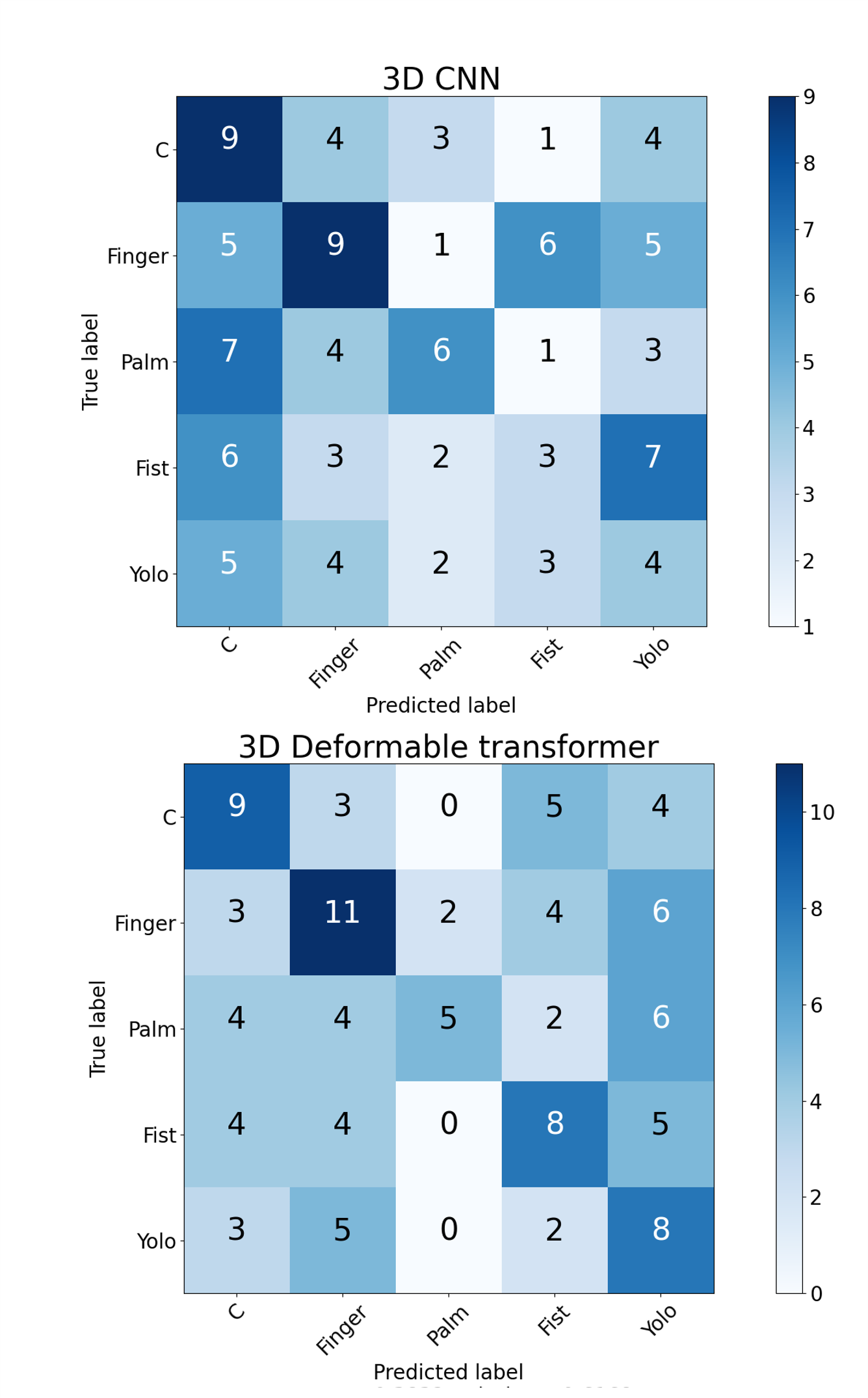}
\caption{Confusion matrices showcasing inter class confusion for (top) 3D CNN and (bottom) 3D Deformable Transformer}
\label{fig:cm}
\end{figure}
\vspace{-\baselineskip}







\section{Conclusion}

In this paper we have demonstrated a radar shape classification using 3D point cloud and deformable transformers as a compelling architecture for classification on radar data. To the best of our knowledge, this is the first time a 3D point-pillar deformable transformer has been used for radar data classification. The 3D transformer model outperforms prior FFT map based convolutional network. The proposed novelties enable powering a wide variety of radar sensing applications without significant model change. For future work, we intend to extend the models for radar object detection as well as temporal activity recognition.


\section*{Acknowledgment}

The authors like to thank Intel Labs for the support of this project.


\bibliographystyle{IEEEtran}

\bibliography{IEEEabrv,IEEEexample}

\end{document}

%% file: EuMW_modify_IEEEtran_18b_CTAN_v4.tex
\makeatletter

\def\@maketitle{\newpage
\bgroup\par\addvspace{0.5\baselineskip}\centering%
\ifCLASSOPTIONtechnote
   {\bfseries\large\@IEEEcompsoconly{\sffamily}\@title\par}\vskip 1.3em{\lineskip .5em\@IEEEcompsoconly{\sffamily}\@author
   \@IEEEspecialpapernotice\par{\@IEEEcompsoconly{\vskip 1.5em\relax
   \@IEEEtitleabstractindextextbox{\@IEEEtitleabstractindextext}\par
   \hfill\@IEEEcompsocdiamondline\hfill\hbox{}\par}}}\relax
\else
   \vskip0.2em{\EuMWtitlesize\ifCLASSOPTIONtransmag\bfseries\LARGE\fi\@IEEEcompsoconly{\sffamily}\@IEEEcompsocconfonly{\normalfont\normalsize\vskip 2\@IEEEnormalsizeunitybaselineskip
   \bfseries\Large}\@title\par}\vskip1.0em\par
   \ifCLASSOPTIONconference%
      {\@IEEEspecialpapernotice\mbox{}\vskip\@IEEEauthorblockconfadjspace%
       \mbox{}\hfill\begin{@IEEEauthorhalign}\@author\end{@IEEEauthorhalign}\hfill\mbox{}\par}\relax
   \else
      \ifCLASSOPTIONpeerreviewca
         {\@IEEEcompsoconly{\sffamily}\@IEEEspecialpapernotice\mbox{}\vskip\@IEEEauthorblockconfadjspace%
          \mbox{}\hfill\begin{@IEEEauthorhalign}\@author\end{@IEEEauthorhalign}\hfill\mbox{}\par
          {\@IEEEcompsoconly{\vskip 1.5em\relax
           \@IEEEtitleabstractindextextbox{\@IEEEtitleabstractindextext}\par\hfill
           \@IEEEcompsocdiamondline\hfill\hbox{}\par}}}\relax
      \else
         \ifCLASSOPTIONtransmag
           {\@IEEEspecialpapernotice\mbox{}\vskip\@IEEEauthorblockconfadjspace%
            \mbox{}\hfill\begin{@IEEEauthorhalign}\@author\end{@IEEEauthorhalign}\hfill\mbox{}\par
           {\vspace{0.5\baselineskip}\relax\@IEEEtitleabstractindextextbox{\@IEEEtitleabstractindextext}\vspace{-1\baselineskip}\par}}\relax
         \else
           {\lineskip.5em\@IEEEcompsoconly{\sffamily}\sublargesize\@author\@IEEEspecialpapernotice\par
           {\@IEEEcompsoconly{\vskip 1.5em\relax
            \@IEEEtitleabstractindextextbox{\@IEEEtitleabstractindextext}\par\hfill
            \@IEEEcompsocdiamondline\hfill\hbox{}\par}}}\relax
         \fi
      \fi
   \fi
\fi\par\addvspace{0.0\baselineskip}\egroup}

\def\EuMWtitlesize{\@setfontsize{\EuMWtitlesize}{24}{24pt}}
\def\EuMWauthorsize{\@setfontsize{\EuMWauthorsize}{11}{11pt}}
\def\EuMWaffilsize{\@setfontsize{\EuMWaffilsize}{10}{10pt}}
\def\EuMWcaptionsize{\@setfontsize{\EuMWcaptionsize}{9}{10pt}}
\def\EuMWbibsize{\@setfontsize{\EuMWbibsize}{8}{10pt}}

\def\@IEEEauthorblockNstyle{\EuMWauthorsize\@IEEEcompsocnotconfonly{\sffamily}\@IEEEcompsocconfonly{\large}}
\def\@IEEEauthorblockAstyle{\EuMWaffilsize\@IEEEcompsocnotconfonly{\sffamily}\@IEEEcompsocconfonly{\itshape}\@IEEEcompsocconfonly{\large}}
\def\@IEEEauthordefaulttextstyle{\EuMWauthorsize\@IEEEcompsocnotconfonly{\sffamily}\sublargesize}

\def\thebibliography#1{\section*{\refname}%
    \addcontentsline{toc}{section}{\refname}%
    \EuMWbibsize\@IEEEcompsocconfonly{\small}\vskip 0.3\baselineskip plus 0.1\baselineskip minus 0.1\baselineskip
    \list{\@biblabel{\@arabic\c@enumiv}}%
    {\settowidth\labelwidth{\@biblabel{#1}}%
    \leftmargin\labelwidth
    \advance\leftmargin\labelsep\relax
    \itemsep \IEEEbibitemsep\relax
    \usecounter{enumiv}%
    \let\p@enumiv\@empty
    \renewcommand\theenumiv{\@arabic\c@enumiv}}%
    \let\@IEEElatexbibitem\bibitem%
    \def\bibitem{\@IEEEbibitemprefix\@IEEElatexbibitem}%
\def\newblock{\hskip .11em plus .33em minus .07em}%
\ifCLASSOPTIONtechnote\sloppy\clubpenalty4000\widowpenalty4000\interlinepenalty100%
\else\sloppy\clubpenalty4000\widowpenalty4000\interlinepenalty500\fi%
    \sfcode`\.=1000\relax}

\long\def\@makecaption#1#2{%
\ifx\@captype\@IEEEtablestring%
\par\@IEEEtabletopskipstrut
\else
\@IEEEfigurecaptionsepspace
\fi
\setbox\@tempboxa\hbox{\normalfont\footnotesize {#1.}\nobreakspace\nobreakspace #2}%
\ifdim \wd\@tempboxa >\hsize%
\setbox\@tempboxa\hbox{\normalfont\footnotesize {#1.}\nobreakspace\nobreakspace}%
\parbox[t]{\hsize}{\normalfont\footnotesize\noindent\unhbox\@tempboxa#2}%
\else
\ifCLASSOPTIONconference \hbox to\hsize{\normalfont\footnotesize\hfil\box\@tempboxa\hfil}%
\else \hbox to\hsize{\normalfont\footnotesize\box\@tempboxa\hfil}%
\fi\fi
\ifx\@captype\@IEEEtablestring%
\@IEEEtablecaptionsepspace
\else
\fi}

\newlength\tablecaptiontotableskip
\newlength\figuretocaptionskip
\def\@IEEEfigurecaptionsepspace{\vskip\figuretocaptionskip\relax}%
\def\@IEEEtablecaptionsepspace{\vskip\tablecaptiontotableskip\relax}%

\def\abstract{\normalfont%
\@IEEEabskeysecsize\bfseries\textit{\abstractname}\,\bfseries\textit{---}\,%
\@IEEEgobbleleadPARNLSP}%

\def\IEEEkeywords{\normalfont%
\@IEEEabskeysecsize\bfseries\textit{\IEEEkeywordsname}\,\bfseries\textit{---}\,%
\@IEEEgobbleleadPARNLSP}%
\def\endIEEEkeywords{\relax\vspace{0.67ex}%
\par\if@twocolumn\else\endquotation\fi%
\normalsize\normalfont}%

\DeclareRobustCommand*{\EuMWauthorrefmark}[1]{\raisebox{0pt}[0pt][0pt]{\textsuperscript{\footnotesize{#1}}}}%
\def\@IEEEauthorblockNtopspace{0ex}
\def\@IEEEauthorblockAtopspace{1mm}
\def\tablename{Table}
\def\thetable{\arabic{table}}
\def\IEEEkeywordsname{Keywords}
\def\subsubsection{\@startsection{subsubsection}{3}{\z@}{1.5ex plus 1.5ex minus 0.5ex}%
{0.7ex plus .5ex minus 0ex}{\normalfont\normalsize\itshape}}%
\newlength{\CPheadmatchindent}%
\def\@seccntformat#1{\hbox to\CPheadmatchindent{\csname the#1dis\endcsname}\hskip 0.1em \relax}
\IEEEilabelindentA \parindent
\IEEEilabelindent \IEEEilabelindentA
\IEEEelabelindent \parindent
\IEEEdlabelindent \parindent
\IEEElabelindent \parindent
\makeatother